\documentclass[10pt,twocolumn,letterpaper]{article}

\usepackage{cvpr}              
\usepackage{times}
\usepackage{epsfig}
\usepackage{graphicx}
\usepackage{amsmath}
\usepackage{amssymb}
\usepackage{array}
\graphicspath{{./Figs/}}
\usepackage[inkscapelatex=false]{svg}
\usepackage{multirow}
\usepackage{soul}
\usepackage{bbding}
\usepackage{lineno}
\usepackage{url}
\usepackage[accsupp]{axessibility}

\definecolor{cvprblue}{rgb}{0.21,0.49,0.74}
\usepackage[pagebackref,breaklinks,colorlinks,citecolor=cvprblue,hyperfootnotes=false]{hyperref}

\def\paperID{2} 
\def\confName{CVPR}
\def\confYear{2024}

\title{AsymFormer: Asymmetrical Cross-Modal Representation Learning for Mobile Platform Real-Time RGB-D Semantic Segmentation}

\author{Siqi Du\textsuperscript{1, $\nmid$}, Weixi Wang\textsuperscript{1, 
 $\nmid$}, Renzhong Guo\textsuperscript{1}, Ruisheng Wang\textsuperscript{1, 2}, Yibin Tian\textsuperscript{1}, Shengjun Tang\textsuperscript{1, *}\vspace {+0.5em}\\ 
\textsuperscript{1}Shenzhen University, China\\
\textsuperscript{2}University of Calgary, Canada\\ 
{\tt\small dusiqi2021@email.szu.edu.cn}\\
{\tt\small \{wangwx, guorz, ybtian, shengjuntang\}@szu.edu.cn}\\
{\tt\small ruiswang@ucalgary.ca}
}

\begin{document}
\maketitle

\begin{abstract}
Understanding indoor scenes is crucial for urban studies. Considering the dynamic nature of indoor environments, effective semantic segmentation requires both real-time operation and high accuracy.To address this, we propose AsymFormer, a novel network that improves real-time semantic segmentation accuracy using RGB-D multi-modal information without substantially increasing network complexity. AsymFormer uses an asymmetrical backbone for multimodal feature extraction, reducing redundant parameters by optimizing computational resource distribution. To fuse asymmetric multimodal features, a Local Attention-Guided Feature Selection (LAFS) module is used to selectively fuse features from different modalities by leveraging their dependencies. Subsequently, a Cross-Modal Attention-Guided Feature Correlation Embedding (CMA) module is introduced to further extract cross-modal representations. The AsymFormer demonstrates competitive results with 54.1\% mIoU on NYUv2 and 49.1\% mIoU on SUNRGBD. Notably, AsymFormer achieves an inference speed of 65 FPS (79 FPS after implementing mixed precision quantization) on RTX3090, demonstrating that AsymFormer can strike a balance between high accuracy and efficiency. \href{https://github.com/Fourier7754/AsymFormer}{Code: \textcolor[RGB]{78,101,155}{https://github.com/Fourier7754/AsymFormer}}

\end{abstract}
\vspace {-1.0em}
\footnote{$\nmid$ indicates equal contribution.}
\footnote{* corresponding author.}
\begin{figure}[htbp]
\centering
\includegraphics[width=0.48\textwidth]{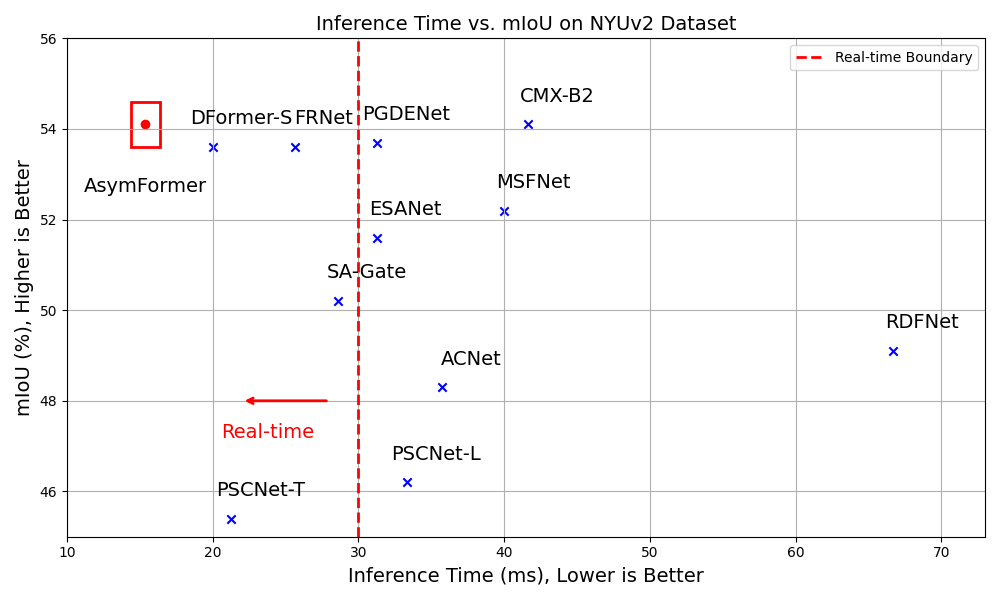}
\caption{\label{fig:散点图} The AsymFormer has 33.0 million parameters and 36.0 GFLOPs computational cost, and it can achieve 65 FPS inference speed on RTX 3090, 54.1\% mIoU on NYUv2.}
\vspace {-0.7em}
\end{figure}

\section{Introduction}
\label{sec:intro}

Indoor scenes are essential to urban environments. Current urban studies necessitate understanding indoor scene semantic information for tasks like emergency evacuation \cite{yoo2022indoor}, robotic navigation \cite{yeboah2018semantic}, and virtual reality \cite{zhang2017physically}. The dynamic nature of indoor scenes demands perception of environmental changes in real time for tasks such as emergency evacuation \cite{yoo2022indoor}, stressing the importance of algorithms' real-time capabilities. Existing real-time semantic segmentation methods often falter when applied to the complex semantic information of indoor scenes, usually requiring a sacrifice in accuracy \cite{yu2021bisenet}. This often requires a trade-off between inference speed and increased network complexity to achieve adequate segmentation accuracy indoors. Consequently, a critical research question is how to improve the accuracy of semantic segmentation in indoor environments without substantially increasing complexity, while ensuring real-time performance.

Aside from increasing the complexity of the network, introducing additional information, such as RGB-D data, is also an effective way to improve the accuracy of semantic segmentation networks. RGB-D cameras are widely used devices for indoor information acquisition. RGB-D information consists of RGB (color, texture and shape) and Depth (boundaries and relative location) features, which are somewhat complementary \cite{zhang2023cmx}. Several studies have explored how to improve indoor scene semantic segmentation performance by integrating RGB-D information \cite{zhang2023cmx,seichter2021efficient,zhou2022frnet}.

Existing research has explored the implementation of attention mechanisms to extract valuable information from RGB-D features without significantly increasing computational complexity. However, due to the additional feature extraction branch for depth features and the lack of discussion on how to allocate computational resources based on feature importance, these methods often introduce a substantial amount of redundant parameters, significantly reducing their computational efficiency \cite{du2022pscnet}.


To address this issue, this paper introduces AsymFormer, a high-performance real-time network for RGB-D semantic segmentation that employs an asymmetric backbone design. This includes a larger parameter backbone for important RGB features and a smaller backbone for the Depth branch. Regarding framework selection, at the same computational complexity, Transformer often achieves higher accuracy but has a slower inference speed compared to CNN \cite{li2022next}. In order to speed up the main branch, this paper employ a hardware-friendly CNN \cite{liu2022convnet} for the RGB branch and a light-weight Transformer \cite{xie2021segformer} for the Depth branch to further compress the parameters. Considering the differences between different modal representation, to effectively select and fuse asymmetric features, this paper proposes a learnable method for feature information compression and constructs a Local Attention Guided Feature Selection (LAFS) module. Additionally, a Cross-Modal Attention (CMA) module is introduced to embed cross-modal information into pixel-wise fused features. Finally, we employ a lightweight MLP-Decoder\cite{xie2021segformer} to decode semantic information from shallow features.


This paper evaluates AsymFormer on two classic indoor scene semantic segmentation datasets: NYUv2 and SUNRGBD. Meanwhile, the inference speed test is also preformed on Nvidia RTX 3090 platform. The AsymFormer achieves 54.1\% mIoU on NYUv2 and 49.1 mIoU on SUNRGBD, with 65 FPS inference speed (79 FPS with mixed precision quantization using the TensorRT). Our experiments highlight AsymFormer's ability to acquire high accuracy and efficiency at the same time. The main contributions are summarized as follows: \begin{itemize} \item[$\bullet$] We employed an asymmetric backbone that compressed the parameters of the Depth feature extraction branch, thus reducing redundancy.

\item[$\bullet$] We introduced the LAFS module for feature selection, utilizing learnable feature weights to calculate spatial attention weights.

\item[$\bullet$] We introduce a novel efficient cross-modal attention (CMA) for modeling of self-similarity in multi-modal features, validating its capability to enhance network accuracy with minimal additional model parameters.

\end{itemize}

\section{Related Works}
\begin{figure*}[htbp]
\centering
\includegraphics[width=0.9\textwidth]{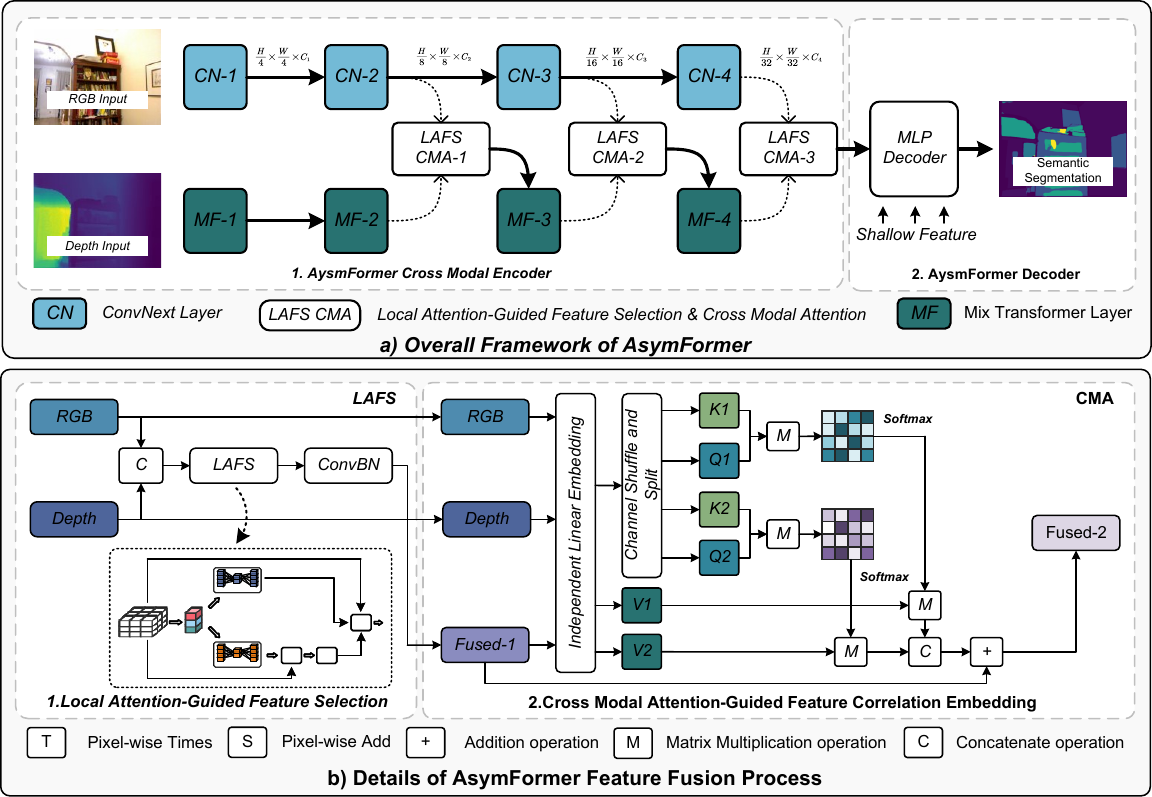}
\caption{\label{fig:overview framework}Overview of AsymFormer.}
\vspace {-1.1em}
\end{figure*}

\label{sec:formatting}
\subsection{Indoor Scene Understanding}
Current research in indoor scene understanding leverages diverse data sources, including RGB images \cite{baruch2021arkitscenes}, RGB-D images \cite{zhang2023cmx}, point clouds \cite{zhou2023bcinet}, and mesh data \cite{zheng2013beyond}. Each of these sources provides unique insights and benefits for analyzing and interpreting indoor environments. For instance, RGB images are accessible and straightforward for visual representations, whereas point clouds and meshes offer intricate 3D spatial data. However, when considering the simultaneous use of both 2D and 3D information to optimize efficiency and effectiveness in scene understanding, RGB-D images emerge as the optimal choice.

\subsection{RGB-D Representation Learning}

One of the earliest works on RGB-D semantic segmentation, FCN \cite{long2015fully}, treated RGB-D information as a single input and processed it with a single backbone. However, subsequent works have recognized the need to extract features from RGB and Depth information separately, as they have different properties. Therefore, most of them have adopted two symmetric backbones for RGB and Depth feature extraction \cite{seichter2021efficient,seichter2022efficient,chen2020bi,zhang2023cmx}. Primarily, asymmetric backbones doubles the overall computational complexity \cite{du2022pscnet}. However, it is generally observed that for semantic segmentation, RGB information typically plays a more prominent role than Depth information, as indicated by \cite{du2022pscnet}. Using a asymmetric backbone for feature extraction will obviously lead to redundant parameters on the less important side, reducing the efficiency of the network.

\subsection{RGB-D feature fusion} 
The performance and efficiency of different frameworks depends largely on how they fuse RGB and Depth features. Some early works, such as RedNet \cite{jiang2018rednet}, fused RGB and Depth feature maps pixel-wise in the backbone. Later, ESANet series \cite{seichter2021efficient, seichter2022efficient} proposed channel attention to select features from different channels, as RGB and Depth feature maps may not align well on the corresponding channels. PSCNet \cite{du2022pscnet} further extended channel attention to both spatial and channel directions and achieved better performance. Recently, more complex models have been proposed to exploit cross-modal information and select features for RGB-D fusion. For example, SAGate \cite{chen2020bi} proposed a gated attention mechanism that can leverage cross-modal information for feature selection. CANet \cite{zhou2022canet} extended non-local attention \cite{wang2018non} to cross-modal semantic information and achieved significant improvement. CMX \cite{zhang2023cmx} extended SA-Gate to spatial and channel directions and proposed a novel cross-modal attention with global receptive field. However, integrating cross-modal information and learning cross-modal similarity is still an open question in vision tasks.

\section{Method}
\subsection{Framework Overview}
This paper develops a high-accuracy real-time semantic segmentation method, AsymFormer, which leverages an asymmetric backbone network to mine multimodal representations. A hardware friendly convolution network ConvNext \cite{liu2022convnet} is used for RGB feature extraction and a light-weight Mix-Transformer \cite{xie2021segformer}is used for processing Depth feature. To effectively fuse RGB-D features, the study introduces a Local Attention Guided Multimodal Feature Selection (LAFS) module, which uses learnable strategy to extract global information and selects multimodal features in both spatial and channel dimensions. Moreover, to further mine the information contained in multimodal features, the study embeds the information contained in multimodal features through a novel Cross-Modal Attention (CMA) module. The overall framework of AsymFormer is shown in Fig.\ref{fig:overview framework}.

\subsection{Local Attention Guided Feature Selection}
Studies have shown that attention mechanisms can effectively select complementary features from RGB and Depth features, thereby improving the efficiency of feature extraction and the performance of semantic segmentation \cite{hu2019acnet,yu2021bisenet,du2022pscnet,zhang2023cmx}. However, existing attention mechanisms usually adopt a non-learnable, fixed strategy for feature information compression \cite{woo2018cbam}, which may overlook the differences among various modal features, causing insufficient use of information.

\begin{figure}[ht]
\vspace {-0.9em}
\centering
\includegraphics[width=0.4\textwidth]{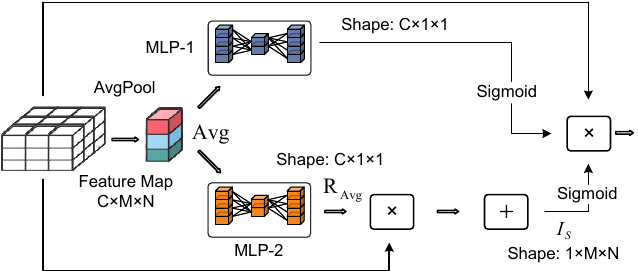}
\caption{\label{fig:LAFS}LAFS.}
\vspace {-0.9em}
\end{figure}

In response to this issue, this paper proposes a learnable method for feature information compression and constructs a Local Attention Guided Feature Selection (LAFS) module. The core design of LAFS is a learnable method for spatial information compression. Inspired by SE \cite{hu2018squeeze} attention, LAFS abandons traditional fixed strategies and employs a feedforward neural network to learn a set of dynamic spatial information compression rules.

The Fig. \ref{fig:LAFS} demonstrates details of the LAFS module. The input features to the LAFS module are the tensor concatenation of $RGB$ and $Depth$ features. For computing attention weights, the study first extracts a global information vector $Avg$ through Adaptive Average Pooling. For channel attention, its structure is the same as SE \cite{hu2018squeeze}.

On spatial attention computation, which is the primary improvement, the LAFS first processes global information $Avg$ through another feedforward neural network with a squeeze-excitation structure, producing the output vector $R_{Avg}$, representing pixel space similarity descriptions. Then, by computing the dot product similarity between $R_{Avg}$ and input feature map, the global spatial information $I_S$ is extracted (this equates to a weighted sum of the channel features, dynamically varying with $Avg$ changes). Subsequently, spatial attention weights $W_{S}$ are calculated through sigmoid normalization.
\begin{equation}
W_{S}=\text{Sigmoid}(\frac{\text{Dot}(Input.\text{Reshape}(C,H \times W)^T,R_{Avg})}{C^{2}})
\end{equation}
Here, all results are divided by $C^{2}$ to prevent sigmoid overflow. Finally, input features are selected based on the pixel-wise multiplication of feature maps with spatial and channel attention weights.

\subsection{Cross-Attention Guided Feature Embedding}
Aside from selecting existing features, it is necessary to mine new information from fused features. Existing Multi-Head Self-Attention (MHSA) in Transformer \cite{vaswani2017attention} is limited to learning self-similarity within a single modality, whereas mining information from multi-modal jointly is a new goal in representation learning. In this paper, a new Cross-Modal Attention (CMA) module is constructed. The key to CMA is defining cross-modal self-similarity using a linear sum, and embedding its result into the fused features.

\subsubsection{Definition of Cross-Modal Self-Similarity:} Suppose $RGB$ and $Depth$ features are seamlessly embedded into $Key$ and $Query$, then for a pixel $(i_0,j_0)$, its cross-modal self-similarity with other pixels $(i,j)$ can be defined as:
\begin{equation}
W(i,j)=\sum_{n=1}^{N}(Kr_{n,i,j} \cdot Qr_{n,i_0,j_0})+\sum_{n=1}^{N}(Kd_{n,i,j} \cdot Qd_{n,i_0,j_0})\
\end{equation}

where $Kr_{n,i,j}$ represents the nth feature value of the pixel $Key_{RGB}$, and $Qr_{n,i,j}$ represents the nth feature value of the pixel $Query_{RGB}$. Similarly, $Kd_{1,i,j}$ and $Qd_{1,i,j}$ represent the nth feature value of the pixel $Key_{Depth}$ and $Query_{Depth}$ respectively.

\subsubsection{Feature Embedding} 
The CMA has three input features, namely $RGB$ features, $Depth$ features, and the fused features $Fused$ selected by LAFS. In the calculation process of CMA, the first step is to embed the input features $RGB$, $Depth$, and $Fused$ into vector space.

\begin{figure}[ht]
\vspace {-0.9em}
\centering
\includegraphics[width=0.4\textwidth]{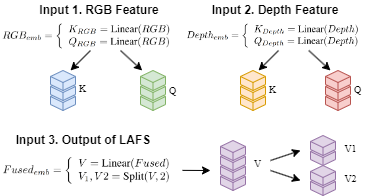}
\caption{\label{fig:嵌入}Feature Embedding.}
\vspace {-0.9em}
\end{figure}
 
Simultaneously, the output $Value$ split into two independent vectors $V_1$ and $V_2$ in the channel direction.

\subsubsection{Splitting and Mixing of Multimodal Information}

To learn features from multiple subspaces, the embedded features $K_{RGB}$, $K_{Depth}$, $Q_{RGB}$, and $Q_{Depth}$ are split into two independent vectors.The study first concatenates $Key_{RGB+Depth}$ with $Query_{RGB+Depth}$ in the channel direction, obtaining $Key$ and $Query$:
\begin{equation}
\begin{aligned}
Key, Query=\left\{ \begin{array}{c}
	Key=\text{Cat}[K_{RGB},K_{Depth}]\\
	Query=\text{Cat}[Q_{RGB},Q_{Depth}]\\
\end{array} \right. 
\end{aligned}
\end{equation} 

It can be noted that if the vectors are directly split and self-similarity is calculated in different subspaces, it might result in the inability to simultaneously include RGB and Depth features in different subspaces. Therefore, we introduce a Shuffle mechanism to ensure each $K1$, $K2$, $Q1$, and $Q2$ contain information from both modalities. As shown in Figure \ref{fig:嵌入}.

\begin{figure}[ht]
\vspace {-0.9em}
\centering
\includegraphics[width=0.4\textwidth]{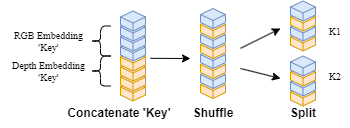}
\caption{\label{fig:混合}Splitting and Mixing of Multimodal Information.}
\vspace {-0.9em}
\end{figure}

After channel shuffle, CMA can ensure that both RGB and Depth information can be utilized simultaneously when calculating self-similarity. 

\subsubsection{Representation Learning in Multiple Subspaces} Finally, CMA calculates the cross-modal similarity and embeds the calculation result into the fused feature $Value$, similar to operations in \cite{vaswani2017attention,wang2018non,dosovitskiy2020image}. Specifically, the study computes the dot products $K_1$, $Q_1$ and $K_2$, $Q_2$ to derive two representations of subspaces $W_1$ and $W_2$:
\begin{equation}
\begin{aligned}
W_1=\text{Softmax}(\frac{Q_1 \cdot K_{1}^T}{\sqrt{C_1/4}})\\
W_2=\text{Softmax}(\frac{Q_2 \cdot K_{2}^T}{\sqrt{C_1/4}})\\
\end{aligned}
\end{equation}
Then, $W_1$ and $W_2$ embed information into $V_1$ and $V_2$ through dot products. The fused feature $Fused_2$ is the concatenation of $V_1$ and $V_2$ in the channel dimension:
\begin{equation}
Fused=\text{Cat}[W_1 \cdot V_1,W_2 \cdot V_2]\\
\end{equation}
Finally, CMA converts $Fused_2$ into output channels $C_2$ using a linear layer and connects it with the residual connection $Fused$, resulting in CMA's output feature.
\label{sec:experiment}
\begin{table*}[ht]
\centering
\caption{\label{tab:ablation experiment}Ablation experiment results for different multi-modal feature fusion method.}
\begin{tabular}{c|cccc|ccc}
\hline \hline
\multirow{2}{*}{\textbf{Model}} & \multicolumn{4}{c|}{\textbf{Feature Fusion Method}} & \multicolumn{3}{c}{\textbf{Metric}} \\ \cline{2-8} 
 & \textbf{Cat} & \multicolumn{1}{c|}{\textbf{SE+MHSA}} & \textbf{LAFS} & \textbf{CMA} & \textbf{Params/M} & \textbf{mIoU(\%)} & \textbf{Inf.Speed(FPS)} \\ \hline
\multirow{2}{*}{\textbf{Baseline}} & \checkmark & \multicolumn{1}{c|}{} &  &  & 31.9 & 47.0 & 77.5 \\
 &  & \multicolumn{1}{c|}{\checkmark} &  &  & 32.6(+0.7M) & 49.9 (+2.9\%) & 65.7 \\ \hline
\multirow{3}{*}{\textbf{Ours}} &  & \multicolumn{1}{c|}{} & \checkmark &  & 32.4 (+0.5M) & 49.1 (+2.1\%) & 75.7 \\
 &  & \multicolumn{1}{c|}{} &  & \checkmark & 32.5 (+0.6M) & 49.6 (+2.6\%) & 67.4 \\
 &  & \multicolumn{1}{c|}{} & \checkmark & \checkmark & 33.0 (+1.1M) & 54.1 (+7.1\%) & 65.5 \\ \hline \hline
\end{tabular}
\vspace {-1.1em}
\end{table*}
\section{EXPERIMENT RESULTS}
\begin{table*}[htbp]
\centering
\vspace {+0.3em}
\caption{\label{tab:NYUv2}Comparison Results on NYUv2. The inference speed is tested on RTX 3090 platform, (480 $\times$ 640) inputs. MS denotes Multi-Scale inference strategy.}
\begin{tabular}{c|ccc|c|c|c|c}
\hline \hline
\textbf{Method} & \textbf{Year} & \textbf{Backbone} & \textbf{Params/M} & \textbf{mIoU (\%)} & \textbf{Real-Time} & \textbf{Speed/FPS} & \textbf{Speed (\%)} \\ \hline
CMX-B2 \cite{zhang2023cmx} & 2022 & Segformer-B2 & 67 & 54.1 & $\times$ & 24 & 36.9\% \\
Token-Fusion \cite{wang2022multimodal} & 2022 & Token-Fusion(S) & - & 54.2 & $\times$ & - & -\\
CMX-B2 (MS) \cite{zhang2023cmx} & 2022 & Segformer-B2 & 67 & 54.4 & $\times$ & - & - \\
Multi-MAE \cite{bachmann2022multimae} & 2022 & Vit-B & - & 56.0 & $\times$ & - & -\\
CMX-B4 (MS) \cite{zhang2023cmx} & 2022 & Segformer-B4 & 140 & 56.3 & $\times$ & - & - \\
Omnivore \cite{girdhar2022omnivore} & 2022 & Swin-L & - & 56.8 & $\times$ & - & -\\
CMX-B5 (MS) \cite{zhang2023cmx} & 2022 & Segformer-B5 & 181 & 56.9 & $\times$ & - & - \\
\hline
SA-Gate \cite{chen2020bi} & 2021 & Res50 & 65 & 50.2 & $\checkmark$ & 35 & 53.8\% \\
ESANet \cite{seichter2021efficient} & 2022 & Res34-Nbt1D & 34 & 51.6 & $\checkmark$ & 32 & 49.2\% \\
PSCNet-L \cite{du2022pscnet} & 2022 & Res50 & 52 & 46.2 & $\checkmark$ & 30 & 46.2\% \\
PSCNet-T \cite{du2022pscnet} & 2022 & Res50 & 40 & 45.4 & $\checkmark$ & 47 & 72.3\% \\
PGDENet \cite{zhou2022pgdenet} & 2022 & Res34 & 101 & 53.7 & $\checkmark$ & 32 & 49.2\% \\
FRNet \cite{zhou2022frnet} & 2022 & Res34 & 86 & 53.6 & $\checkmark$ & 39 & 60.0\% \\
DFormer-S \cite{yin2023dformer} & 2023 & DFormer-S & 18.7 & 53.6 & $\checkmark$ & 50 & 76.9\% \\ \hline
AsymFormer & 2024 & B0+T & 33 & 54.1 & $\checkmark$ & 65 & 100.0\% \\
AsymFormer (FP16) & 2024 & B0+T & 33 & 54.1 & $\checkmark$ & 79 & 121.5\%\\
AsymFormer (MS) & 2024 & B0+T & 33 & 55.3 & $\times$ & - & -
\\ \hline \hline
\end{tabular}
\vspace {-0.7em}
\end{table*}

\subsection{Implementation Details}
To evaluate our Real-Time semantic segmentation network design, we conduct a series of experiments on two widely-used datasets NYUv2\cite{silberman2012indoor}  (795 training and 654 testing RGB-D images) and SUNRGBD\cite{song2015sun} (5825 training and 5050 testing RGB-D images). We conduct the model training and testing on different platforms. For the training, we use Nvidia A100-40G GPU. For the evaluation and inference speed testing, we use Nvidia RTX 3090 GPU, Ubuntu 20.04, CUDA 12.0 and Pytorch 2.0.1. We apply data augmentation to all datasets by randomly flipping (p=0.5), random scales between 1.0 and 2.0, random crop 480$\times$640 and random HSV. We adopt ConvNext-T \cite{liu2022convnet} backbones pretrained on ImageNet-1k\cite{deng2009imagenet}. For the Mix Transformer-B0 \cite{xie2021segformer}, no pre-trained weight has been used. The MLP-decoder in AsymFormer has the same structure as Segformer and an embedding dimension of 256. We choose AdamW optimizer with a weight decay of 0.01. The initial learning rate is $5e^{-5}$ and we use a poly learning rate schedule $(1-\frac{iter}{max_iter})^{0.9}$ with a warm-up of 10 epochs. We train with a batch size of 8 for NYUv2 (500 epochs) and SUNRGBD (200 epochs). We employ cross-entropy as the loss function and do not use any auxiliary loss during training process. The evaluation metric is mean Intersection over Union (mIoU).


\subsection{Ablation Experiment}
We conduct a series of ablation experiments on NYUv2 dataset to evaluate the effectiveness of the LAFS and CMA module. We set two common feature fusion methods as our comparative baseline: \textbf{1. Cat:} This method directly concatenates two features and then uses convolution layers to adjust the channel numbers. Essentially, it is a pixel-wise fusion without feature selection. \textbf{2. SE+MHSA:} This method combines the popular SE attention \cite{hu2018squeeze} and MHSA attention \cite{vaswani2017attention} for feature fusion. Here, SE is used for feature selection in the channel direction, while MHSA is employed for further feature extraction on the fused features.

In our experiments, the Cat fusion method, used as a baseline, achieved a segmentation accuracy of 47.0 mIoU and an inference speed of 77.5 FPS. When using LAFS alone, we achieved a performance improvement of 2.1\% while sacrificing only 1.8 FPS of inference speed. This demonstrates that LAFS provides performance gains without significantly impacting inference speed. In comparison to the other baseline, using Cat+MHSA, which resulted in a reduction of inference speed by 11.8 FPS, an improvement of only 2.9\% in mIoU was achieved. This further highlights the efficiency of LAFS. Furthermore, when using CMA alone, we observed a 2.6\% improvement in segmentation accuracy but encountered a significant decrease in inference speed of 10.1 FPS. Compared to LAFS, CMA showed a more noticeable reduction in inference speed. 

Finally, we combined LAFS with CMA (LAFS+CMA). Since LAFS had minimal impact on inference speed and served a different purpose than CMA, the network's inference speed decreased by only 2 FPS. This change achieved a significant improvement of 7.1\% in segmentation accuracy compared to the baseline Cat. At this point, the inference speed of LAFS+CMA was similar to SE+MHSA, but with a 4.2\% performance improvement. This validates our experimental hypothesis: by re-modeling feature selection and mining cross-modal self-similarity, we can enhance the segmentation performance of the network without sacrificing inference speed compared to existing models. This demonstrates that we have indeed improved the efficiency of the network.


\subsection{Comparison With State-of-The-Arts}
\subsubsection{NYUv2 Comparison Results}
According to Table \ref{tab:NYUv2}, despite the lack of ImageNet-1k pretraining—a common practice among competing methods—our AsymFormer still achieves leading scores in Real-Time semantic segmentation. Th AsymFormer achieves 54.1 $\%$ mIoU, demonstrating competitive accuracy compared to those high-performance heavy designs. AsymFormer also has faster inference speed than other methods. For instance, AsymFormer outperforms PSCNet-T\cite{du2022pscnet} by 8.7\% mIoU and 18 FPS inference speed improvement. Similarly, AsymFormer is two times faster than ESANet\cite{seichter2021efficient} and three times faster than CMX-B2 \cite{zhang2023cmx} with the same performance. Finally, by using multi-scale inference strategy, the AsymFormer achieves 55.3 \% mIoU on NYUv2. In terms of semantic segmentation accuracy, AsymFormer does not show a significant disadvantage compared to those high-performance methods, such as Omnivore, included in the comparison. This validates the effectiveness of our various efforts in reducing network redundancy parameters and improving inference speed.

\subsubsection{SUNRGBD Comparison Results}
Table \ref{tab:SUNRGBD} reports the performance of AsymFormer on the SUNRGBD dataset. AsymFormer achieves competitive accuracy with 49.1 \% mIoU. The advantage of AsymFormer is not as significant as in NYUv2 experiment. For example, AsymFormer improves 3.9 mIoU over SA-Gate\cite{chen2020bi} in NYUv2 dataset (54.1\% vs 50.2\% mIoU), but decreases 0.3 mIoU in SUNRGBD dataset (49.1\% vs 49.4\% mIoU). A similar performance degradation can be observed in CMX-B2 result which also uses Transformer based backbone. We conjecture that this phenomenon may be caused by low quality depth images in SUNRGBD dataset. The aim of our research is not to construct a state-of-the-art method that has a marginal mIoU improvement over other methods, but to construct a method that has a better performance-speed balance and is more suitable for robot platform. Given that AsymFormer still has faster inference speed than other methods, we consider this performance acceptable for AsymFormer.

\begin{table}[htbp]
\centering
\setlength{\tabcolsep}{3mm}{
\caption{\label{tab:SUNRGBD}Comparison Results on SUNRGBD. MS denotes Multi-Scale inference strategy.}
\begin{tabular}{c|cc}
\hline \hline
\textbf{Method} & \textbf{Pixel Acc. (\%)} & \textbf{mIoU (\%)} \\ \hline 
RDFNet \cite{park2017rdfnet}          & 81.5                     & 47.7               \\
ESANet \cite{seichter2021efficient}         & -                        & 48.0                 \\
ACNet \cite{hu2019acnet}           & -                        & 48.1               \\
SA-Gate \cite{chen2020bi}         & 82.5                     & 49.4               \\
CMX-B2 (MS) \cite{zhang2023cmx}         & 82.8                     & 49.7               \\
DFormer-S \cite{yin2023dformer}         & -                     & 50.0 \\
MSFNet \cite{jiang2022multi}          & -                        & 50.3               \\
FRNet \cite{zhou2022frnet}           & 87.4                     & 51.8               \\
PGDENet \cite{zhou2022pgdenet}         & 87.7                     & 51.0\\
CMX-B4 (MS) \cite{zhang2023cmx}         & 83.5                     & 52.1               \\
CMX-B5 (MS) \cite{zhang2023cmx}         & 83.8                     & 52.4               \\
\hline
AsymFormer   & 81.9                     & 49.1               \\\hline \hline
\end{tabular}}
\vspace {-0.9em}
\end{table}

\subsection{Visualization}
\subsubsection{LAFS Attention Map}

\begin{figure}[ht]
\vspace {-0.9em}
\centering
\includegraphics[width=0.35\textwidth]{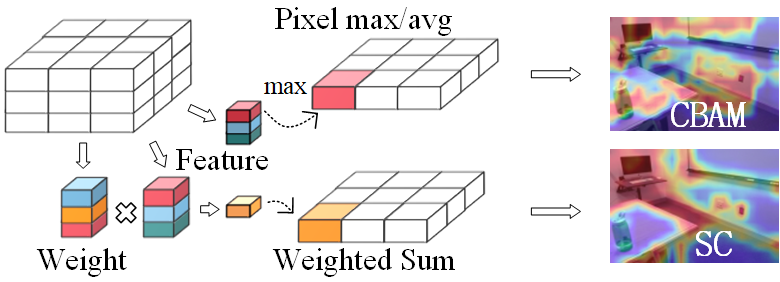}
\caption{\label{fig:CBAM}Difference between CBAM and LAFS.}
\vspace {-0.9em}
\end{figure}

As shown in Figure \ref{fig:CBAM}, to demonstrate that LAFS performs better than CBAM \cite{woo2018cbam} in selecting features in the spatial dimension, we visualized the spatial attention weights of both methods. It can be observed that LAFS provides better coverage of informative regions in the image while maintaining consistency within objects and preserving the integrity of edges.

\subsubsection{Semantic Segmentation Results}
\begin{figure}[htbp]
\centering
\includegraphics[width=0.35\textwidth]{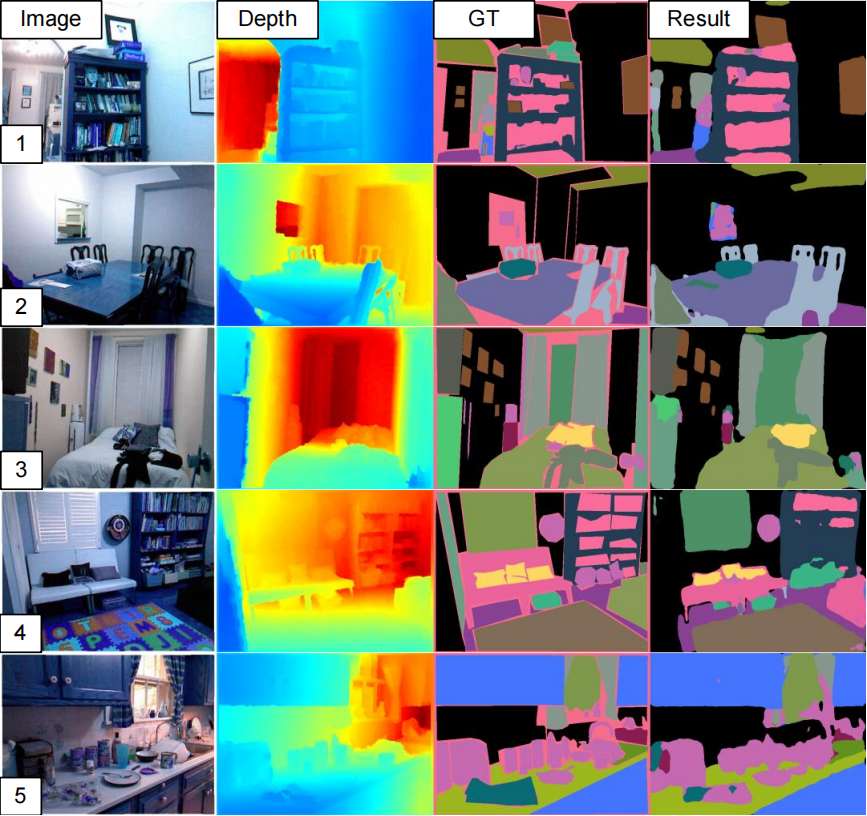}
\caption{\label{fig:result}Visualization of AsymFormer Semantic Segmentation Results.}
\vspace {-1.1em}
\end{figure}

The Figure \ref{fig:result} demonstrates the segmentation results of AsymFormer on the NYUv2 dataset. As observed, while maintaining a significantly faster inference speed compared to other methods, AsymFormer achieves comparable semantic segmentation accuracy to mainstream approaches.

\section{CONCLUSIONS}
\label{sec:conclusion}
In this work, we proposed AsymFormer, which aims to construct a less-redundant real-time indoor scene understanding network. To enhance efficiency and reduce redundant parameters, we implemented the following improvement: 1. the asymmetric backbone that compressed the parameters of the Depth feature extraction branch, thus reducing redundancy. 2. the LAFS module for feature selection, utilizing learnable strategy for global information compressing and improving spatial attention calculations 3. the self-similarity in multi-modal features, validating its capability to enhance network accuracy with minimal additional model parameters. The experiments demonstrated that the AsymFormer achieves a balance between accuracy and speed. Moving forward, we will continue to optimize the modules and address issues such as self-supervised pre-training of the model, aiming for further improvements.
{
    \small
    \bibliographystyle{ieeenat_fullname}
    \bibliography{main}
}


\end{document}